\documentclass[draft]{agujournal2019}
\usepackage{url}
\usepackage{lineno}
\usepackage[inline]{trackchanges}
\usepackage{soul}
\usepackage{graphicx}
\usepackage{amsmath}
\usepackage{amssymb}
\draftfalse

\journalname{JGR Water Resources Research}
\makeatletter
\AtBeginDocument{%
  \let\@oddhead\@empty
  \let\@evenhead\@empty
}
\makeatother
\begin{document}

\title{Ranking Competing geologic interpretations via foundation-model-assisted generative hydrologic inversion}

\authors{H. U. Rashid\affil{1} and D. O'Malley\affil{1}}

\affiliation{1}{Earth and Environmental Sciences Division, Los Alamos National Laboratory, Los Alamos, NM, USA}

\correspondingauthor{H. U. Rashid}{hrashid.lsu@gmail.com}

\begin{keypoints}
    \item We convert geologic interpretations into geologic maps that can be tested against flow observations.
    \item Our framework ranks competing geologic interpretations based on their consistency with limited observations.
    \item Our framework consistently ranks competing interpretations in real-field case studies.
\end{keypoints}

\begin{abstract}
High-consequence subsurface decisions often rely on sparse data that permit competing geological interpretations. Determining consistency of these interpretations with the available observations remains challenging. We present a workflow that addresses this challenge by translating competing geologic interpretations into alternative priors and ranking them according to their consistency with hydraulic-head observations. A key step in this workflow is exploiting the broad knowledge of image-generation foundation models to transform nuanced geologic interpretations into data ready for computer modeling. For each interpretation, a text-to-image foundation model generates an ensemble of geologic images, and a separately trained variational autoencoder learns an interpretation-specific latent representation. A supervised inverse network maps head observations into this latent space, and the frozen decoder reconstructs an image that is mapped to a log-conductivity field. Steady-state flow simulations predict heads, and the aggregate normalized head error determines the ranking. We evaluate the framework using a synthetic benchmark based on the Johansen Formation with three interpretations of decreasing consistency with the reference geology. Across 595 test cases, the Precise \& Accurate interpretation produces lower normalized errors than Accurate in 58.5\% of cases and Mismatched in 82.5\% of cases. Accurate outperforms Mismatched in 65.5\% of cases. We then compare spatial representations of two published conceptual models of the Culebra Dolomite Member at the Waste Isolation Pilot Plant. The revised representation yields an aggregate normalized error of 7.598, compared with 8.595 for the original, consistent with the documented conceptual-model revision. The framework enables quantitative comparison of competing geological interpretations using available hydraulic observations.
\end{abstract}

\section*{Plain Language Summary}
Decisions about the deep subsurface often have to be made when very little is known about the site. Geologists studying the same place frequently disagree about how it is put together, and each records their view as a written description. There is rarely a practical way to tell which description is closest to reality, and the disagreement can persist even after gathering site data. We built a method that tests written descriptions against measurements that already exist. Artificial intelligence tools that generate images from text turn each description into a set of possible maps showing how easily water moves through the rock. A computer model predicts the water pressures each map would produce, and those predictions are compared with pressures measured in wells. The description whose predictions come closest is ranked highest, and the ranking carries a number showing by how much. In tests where the correct answer was known, the method found it. Applied to the Waste Isolation Pilot Plant in New Mexico, where the operator revised its understanding of a key water-bearing layer after new field data arrived, the method independently favored the revised description.

\section{Introduction}
Predictive subsurface models underpin decisions in groundwater management, hydrocarbon production, geothermal development, geologic carbon storage, and subsurface waste isolation. A central component of these models is the geological conceptualization: an interpretation of which units and structures are present, how they are connected, and which spatial features control fluid flow and transport. These interpretations are constructed from incomplete observations such as cores, well logs, seismic and other geophysical data, outcrop analogues, and geological reasoning. Because the subsurface is only sparsely observed, the resulting geological interpretation is inherently nonunique \cite{caers2011,wellmann2018}. In a controlled experiment illustrating this ambiguity, \citeA{bond2007} analyzed 412 interpretations of the same synthetic seismic section and found substantial variability among interpreters; only 21\% identified the intended interpretation. Thus, alternative geological interpretations can arise even when investigators are presented with the same underlying observations.

The consequences of this interpretational uncertainty extend across subsurface applications. In groundwater modeling, an incorrect conceptualization may produce unreliable predictions \cite{bredehoeft2005}, and alternative hydrogeological conceptual models can produce substantially different forecasts \cite{hojberg2005,refsgaard2006,refsgaard2012}. Similar problems arise in petroleum reservoir modeling, where uncertainty in the geological scenario can strongly influence net present value and development decisions \cite{demyanov2019}. In geologic carbon storage, alternative representations of heterogeneity, stratigraphic architecture, and boundary conditions can alter predicted pressure evolution, plume migration, and storage performance \cite{li2011}. Geological uncertainty is likewise central to geothermal reservoir characterization and long-term assessments of subsurface waste isolation \cite{scheidt2018,wellmann2018}. Across these applications, parameter uncertainty within a single geological model therefore represents only part of the total predictive uncertainty.

One strategy for addressing this problem is to construct multiple plausible geological conceptual models and propagate each through the forward model. Their predictions can then be compared or combined using multimodel inference, Bayesian model averaging, or scenario-based uncertainty quantification \cite{neuman2003,ye2004,demyanov2019}. Such approaches account for alternative interpretations once numerical realizations of those interpretations are available. A practical difficulty occurs earlier in the workflow: each geological interpretation must first be translated into a simulation-ready representation. Geological descriptions are commonly expressed through maps, cross sections, conceptual diagrams, and written interpretations, whereas numerical simulators require explicit spatial distributions of facies or physical properties. Constructing these representations for many competing interpretations can require substantial manual effort and may introduce additional subjective choices during model construction \cite{wellmann2018}. 

Dynamic observations such as pressure and hydraulic head provide an integrated test of geologic structure because they reflect the combined effects of unit geometry, connectivity, and property contrasts along flow paths. However, these observations are commonly used to estimate parameters within a geological prior selected in advance rather than to evaluate the interpretation defining that prior. Multimodel approaches can compare alternative interpretations, but only after each has been translated into a numerical representation \cite{refsgaard2012,demyanov2019}. Establishing an efficient link between written interpretations, alternative spatial priors, and common dynamic observations would therefore enable earlier and more systematic evaluation of geological uncertainty.

Generative methods provide a potential means of reducing this barrier. Deep generative models can represent complex geological property fields using low-dimensional latent variables while preserving spatial structures characteristic of the training data. \citeA{laloy2017} used a deep generative representation of complex binary geological media for inverse modeling, while \citeA{laloy2018} developed a spatial generative adversarial network for training-image-based geostatistical inversion. \citeA{mo2020} combined an adversarial autoencoder with a convolutional surrogate for estimating non-Gaussian conductivity fields. More broadly, training-image-based geostatistical methods generate ensembles that reproduce geological patterns specified through representative training images \cite{strebelle2002,mariethoz2014}. These approaches enable geologically realistic parameterizations, but the geological patterns to be represented must still be specified beforehand through training data or training images. The choice of those patterns therefore encodes an important part of the assumed geological conceptualization.

Text-conditioned generative models available via APIs from large technology companies like OpenAI and Google offer a different route for connecting geological interpretation to spatial representation. Latent diffusion models, for example, can generate images conditioned on natural-language descriptions \cite{rombach2022}. This capability creates the possibility of using written geological interpretations directly as conditioning information: alternative descriptions can generate alternative ensembles of geological maps, which can subsequently be tested against dynamic observations. Recent work has also introduced large language models into process-based subsurface modeling. \citeA{ma2026}, for example, developed a multi-agent framework that generates and iteratively modifies calibration code around simulators including MODFLOW-2005 and TOUGHREACT. Such approaches automate components of model implementation and calibration for a specified model structure. Here, we address a complementary problem: whether alternative geological interpretations themselves can be translated into spatial models and evaluated against observations.

This study introduces a framework that treats competing written geologic interpretations as testable hypotheses. The framework converts each interpretation into an ensemble of spatial geologic realizations, conditions an interpretation-specific latent model on sparse hydraulic-head observations, and uses forward-model likelihoods to quantify the relative support for each interpretation within the candidate set. We first evaluate the framework using a controlled benchmark based on the Johansen Formation, comprising three interpretations with progressively decreasing fidelity to the reference geology. We then apply it to two published conceptual models of the Culebra Dolomite Member at the Waste Isolation Pilot Plant (WIPP), where a documented conceptual-model revision provides an independent basis for evaluating the ranking. We show that limited hydraulic observations often contain sufficient information to distinguish among competing geologic interpretations.

The remainder of this paper is organized as follows. Section~2 presents the proposed framework, including the generation of interpretation-specific image ensembles, latent-space representation and inversion, forward flow simulation, and likelihood-based ranking. Section~3 evaluates the framework using a controlled synthetic benchmark comprising three geologic interpretation classes. Section~4 applies the framework to two competing conceptual models of the Culebra Dolomite at WIPP. Section~5 discusses the interpretation of the rankings, the information provided by existing observations, connections to related machine-learning approaches, and the limitations of the framework. Section~6 summarizes the principal findings and identifies priorities for future research.


\section{Method}

\subsection{Problem formulation and workflow}

Let $\{C_1,\ldots,C_K\}$ denote a finite set of candidate geologic interpretations, each expressed as a written description. For observation case $i$, hydraulic heads $\mathbf{d}_i \in \mathbb{R}^{N_{\mathrm{obs}}}$ are available at a fixed set of monitoring locations. The objective is to evaluate how consistently each candidate interpretation can reproduce these observations after being converted into an explicit spatial conductivity model.

\begin{figure}[!h]
\centering
\includegraphics[width=\textwidth]{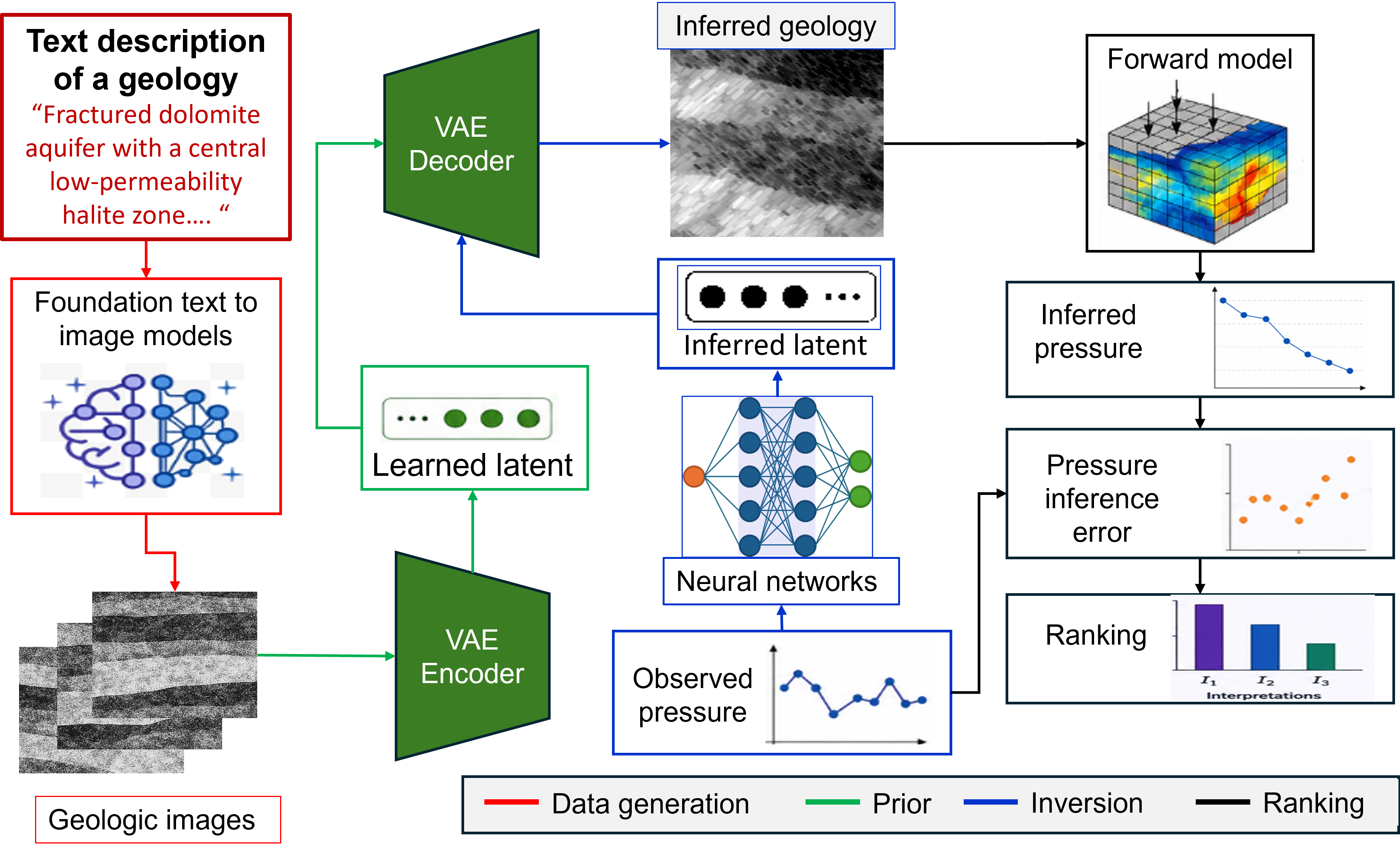}
\caption{Workflow for comparing competing geologic interpretations. A written interpretation is expanded into an image ensemble by a text-conditioned foundation model (red). An interpretation-specific variational autoencoder learns a low-dimensional representation of that ensemble (green). A supervised inverse network maps hydraulic-head observations to a latent vector, and the frozen decoder maps that vector to a conductivity field (blue). Forward simulation produces predicted heads whose mismatch with the observations determines the interpretation ranking (black).}
\label{fig:workflow}
\end{figure}

The workflow contains four stages (Figure~\ref{fig:workflow}). First, a text-conditioned foundation model converts each written interpretation into an ensemble of geologic images. Second, a variational autoencoder (VAE) learns an interpretation-specific low-dimensional representation of that ensemble. Third, a supervised inverse network maps the head observations to the latent space of each interpretation. Finally, the frozen decoder converts the inferred latent vector to a conductivity field, a forward flow simulation predicts heads at the monitoring locations, and the resulting residual is used to rank the candidate interpretations.

For each candidate $C_k$, $f_k:\mathbb{R}^{N_{\mathrm{obs}}}\rightarrow
\mathbb{R}^{d_z}$ denotes the interpretation-specific inverse network.
The vector $\hat{\mathbf{z}}_{ik}\in\mathbb{R}^{d_z}$, where
$d_z=128$, is the latent representation predicted by $f_k$ from the
observation vector $\mathbf{d}_i$. The decoder $D_k$ maps
$\hat{\mathbf{z}}_{ik}$ to a normalized geologic image associated with
candidate $C_k$, and $\mathcal{M}$ maps the image intensities to the
inferred cellwise natural log-conductivity field
$\hat{\mathbf{Y}}_{ik}$, where $\mathbf{Y}=\ln\mathbf{K}$. The forward
operator $\mathcal{F}$ solves the flow equation and produces the complete
simulated head field
$\hat{\mathbf{h}}_{ik}\in\mathbb{R}^{N_{\mathrm{grid}}}$, where
$N_{\mathrm{grid}}=16{,}384$. Finally, the fixed observation operator
$\mathcal{H}:\mathbb{R}^{N_{\mathrm{grid}}}\rightarrow
\mathbb{R}^{N_{\mathrm{obs}}}$ extracts the simulated heads at the
monitoring locations, producing the predicted observation vector
$\hat{\mathbf{d}}_{ik}$. The complete prediction sequence is

\begin{linenomath*}
\begin{equation}
\begin{aligned}
\hat{\mathbf{z}}_{ik}
    &= f_k(\mathbf{d}_i), \\
\hat{\mathbf{Y}}_{ik}
    &= \mathcal{M}\!\left[
       D_k(\hat{\mathbf{z}}_{ik})
       \right], \\
\hat{\mathbf{h}}_{ik}
    &= \mathcal{F}(\hat{\mathbf{Y}}_{ik}), \\
\hat{\mathbf{d}}_{ik}
    &= \mathcal{H}(\hat{\mathbf{h}}_{ik}).
\end{aligned}
\label{eq:workflow_prediction}
\end{equation}
\end{linenomath*}

Here, a hat denotes an inferred or predicted quantity. Because $f_k$ and $D_k$ are trained separately for each candidate interpretation, $\hat{\mathbf{Y}}_{ik}$ is restricted to the spatial patterns represented by the image ensemble associated with $C_k$. The same observation vector $\mathbf{d}_i$ and observation operator $\mathcal{H}$ are used for all candidate interpretations, ensuring that differences among the predicted observations $\hat{\mathbf{d}}_{ik}$ reflect differences among the interpretation-specific models.


\subsection{From interpretation to image ensemble}

For each candidate interpretation $C_k$, the written geological description was submitted as a conditioning prompt through ChatGPT Enterprise \cite{openai2026chatgpt}, with GPT-5.5 selected as the conversational model and using the built-in Image Generation tool. For each interpretation, the model produces an ensemble $\mathcal{X}_k=\{\mathbf{x}_{kn}\}_{n=1}^{N_k}$ of grayscale images with a resolution of $256\times256$ pixels. Pixel intensities are represented on $[0,1]$ and are treated as normalized indicators of spatial variation in log conductivity: larger values correspond to more conductive regions. The generated images are not calibrated property fields. Rather, they represent spatial characteristics conveyed by the interpretation, including layering, contact geometry, continuity, and the presence or absence of connected high-conductivity features. The conversion to numerical conductivity values is imposed separately through $\mathcal{M}$.

Written descriptions convey depositional setting and qualitative spatial relations more readily than detailed geometry. Images can also be used to augment the text. For example, in applications with an irregular external boundary or a geometrically complex internal contact that is known but hard to describe in words, an outline image can therefore be supplied as an additional conditioning input. The text then specifies the interpreted internal architecture, whereas the outline constrains its location and geometry. This auxiliary conditioning is used for the WIPP application described in Section~4.2.

\subsection{Interpretation-specific latent representation}

One convolutional VAE \cite{kingma2014} is trained independently for each interpretation using only the corresponding image ensemble. Our encoder contains four convolutional layers with $4\times4$ kernels, stride 2, unit padding, and 32, 64, 128, and 256 output channels, respectively. Each convolution is followed by a ReLU activation. For a $256\times256$ input image, the resulting $16\times16\times256$ feature array is flattened and passed to two linear layers that return the mean $\boldsymbol{\mu}_k(\mathbf{x})$ and log variance $\boldsymbol{\ell}_k(\mathbf{x})$ of a $d_z=128$ dimensional (approximately) Gaussian latent distribution. During training, a latent sample is obtained using the reparameterization
\begin{linenomath*}
\begin{equation}
\mathbf{z}=\boldsymbol{\mu}_k+\exp\!\left(\tfrac{1}{2}\boldsymbol{\ell}_k\right)\odot\boldsymbol{\epsilon},
\qquad \boldsymbol{\epsilon}\sim\mathcal{N}(\mathbf{0},\mathbf{I}).
\label{eq:reparameterization}
\end{equation}
\end{linenomath*}
Our decoder applies a linear layer followed by four transposed-convolution layers that reverse the encoder dimensions; the final sigmoid activation constrains reconstructed intensities to $[0,1]$.

The VAE objective function is
\begin{linenomath*}
\begin{equation}
\mathcal{L}_{\mathrm{VAE}}
= \operatorname{MAE}(\hat{\mathbf{x}},\mathbf{x})
+ \lambda_e E(\hat{\mathbf{x}},\mathbf{x})
+ \beta(t)D_{\mathrm{KL}}\!\left[q_k(\mathbf{z}\mid\mathbf{x})\,\Vert\,\mathcal{N}(\mathbf{0},\mathbf{I})\right],
\label{eq:vae}
\end{equation}
\end{linenomath*}

where $\operatorname{MAE}$ denotes the mean absolute error.
The edge term compares local intensity differences between
the reconstructed and input images:
\begin{linenomath*}
\begin{equation}
\begin{aligned}
E(\hat{\mathbf{x}},\mathbf{x})
&= \operatorname{MAE}
   (\Delta_{1}\hat{\mathbf{x}},\Delta_{1}\mathbf{x}) \\
&\quad + \operatorname{MAE}
   (\Delta_{2}\hat{\mathbf{x}},\Delta_{2}\mathbf{x}).
\end{aligned}
\label{eq:edge}
\end{equation}
\end{linenomath*}

Here, $\Delta_{1}$ and $\Delta_{2}$ are first-order
forward-difference operators along the row and column
directions, respectively. For an image with $H$ rows
and $W$ columns, they are defined as
\begin{linenomath*}
\begin{equation}
\begin{aligned}
(\Delta_{1}\mathbf{x})_{r,c}
&= x_{r+1,c}-x_{r,c}, \\
(\Delta_{2}\mathbf{x})_{r,c}
&= x_{r,c+1}-x_{r,c}.
\end{aligned}
\label{eq:image_differences}
\end{equation}
\end{linenomath*}

Here, $x_{r,c}$ denotes the intensity at row $r$ and
column $c$. For $\Delta_{1}$, the index ranges are
$r=1,\ldots,H-1$ and $c=1,\ldots,W$.
For $\Delta_{2}$, they are $r=1,\ldots,H$ and
$c=1,\ldots,W-1$.
The same operators are applied to $\hat{\mathbf{x}}$.
Differences use unit pixel spacing and are evaluated
only between neighboring pixels within the image,
without boundary padding. Each MAE averages the
mismatch over the corresponding valid pixel pairs.
This term penalizes errors in local intensity contrasts,
encouraging the preservation of geologic contacts and
limiting excessive smoothing during reconstruction.
We set $\lambda_{e}=0.1$. The Kullback--Leibler weight $\beta(t)$ increases linearly from 0 to 0.01 over the first 100 epochs and remains at 0.01 thereafter.

The VAEs are trained for 3000 epochs with Adam, a learning rate of $10^{-4}$, and a batch size of 32. Each image ensemble is divided randomly into 80\% training and 20\% validation subsets. Training uses stochastic latent samples, whereas validation decodes the posterior mean $\boldsymbol{\mu}_k$ to remove sampling variability. The retained checkpoint is the one with the smallest deterministic validation reconstruction loss, defined as the sum of the pixel MAE and the weighted edge loss; the Kullback--Leibler term is not used in checkpoint selection.

\subsection{Flow model and observation operator}

Forward simulations are performed with DPFEHM \cite{omalley2023}, a differentiable subsurface-flow package implemented in Julia \cite{bezanson2017} and used here with Flux \cite{innes2018}. The synthetic benchmark solves two-dimensional, steady-state, single-phase flow,
\begin{linenomath*}
\begin{equation}
\nabla\!\cdot\!\left[K(\mathbf{s})\nabla h(\mathbf{s})\right]+q(\mathbf{s})=0,
\label{eq:flow}
\end{equation}
\end{linenomath*}
where $K$ is the hydraulic conductivity, $h$ is the hydraulic head, $q$ is fluid sources/sinks, and $\mathbf{s}$ is a position in space. The equation is solved on a regular $128\times128$ grid spanning $[-100,100]\,\mathrm{m}$ in both horizontal directions and having unit thickness. No internal source or sink is applied, so $q=0$.

A decoded image is sampled at the flow-grid locations and mapped to natural log conductivity,
\begin{linenomath*}
\begin{equation}
Y=\ln K=Y_{\min}+g(Y_{\max}-Y_{\min}),
\qquad Y_{\min}=-30, \qquad Y_{\max}=-25.
\label{eq:logk}
\end{equation}
\end{linenomath*}
Thus, the image range spans five natural-log units, corresponding to a conductivity ratio of $\exp(5)\approx148$ between the endpoints ($g\in[0,1]$ is from the image generated by the foundation model). Conductivity on the face between adjacent cells $a$ and $b$ is evaluated using the geometric mean,
\begin{linenomath*}
\begin{equation}
K_{ab}=\exp\!\left(\frac{Y_a+Y_b}{2}\right).
\label{eq:facek}
\end{equation}
\end{linenomath*}

Prescribed heads impose a lateral gradient: $h=10\,\mathrm{m}$ at $x=-100\,\mathrm{m}$ and $h=0\,\mathrm{m}$ at $x=100\,\mathrm{m}$. We use a  set of $N_{\mathrm{obs}}=250$ monitoring nodes and keep them fixed for inverse-model training and comparison.

For a synthetic field with clean monitored-head vector $\mathbf{d}^{0}_i$, observations are generated using
\begin{linenomath*}
\begin{equation}
\mathbf{d}_i=\mathbf{d}^{0}_i+\boldsymbol{\eta}_i,
\qquad
\boldsymbol{\eta}_i\sim\mathcal{N}(\mathbf{0},\sigma_{\eta,i}^{2}\mathbf{I}),
\qquad
\sigma_{\eta,i}=0.01\,\overline{|\mathbf{d}^{0}_i|},
\label{eq:observation_noise}
\end{equation}
\end{linenomath*}
where $\overline{|\mathbf{d}^{0}_i|}$ is the mean absolute clean head. The same noisy observation vector is supplied to every candidate interpretation for a given test field. The synthetic comparison uses the noisy observations as the reference for ranking.

\subsection{Supervised latent inverse operator}

An inverse network $f_k$ is trained separately for each interpretation. Because the monitoring locations are fixed, their heads are represented as an ordered vector. The network contains three fully connected hidden layers with 256 units and ReLU activations, followed by a linear layer that maps $\mathbb{R}^{250}$ to the 128-dimensional latent space.

For every training image $\mathbf{x}$, the frozen VAE encoder provides the target latent mean $\boldsymbol{\mu}_k(\mathbf{x})$. The original image, rather than its VAE reconstruction, is mapped through Equation~\ref{eq:logk} to obtain the target field $\mathbf{Y}$. A forward simulation of that field supplies the clean head vector $\mathbf{d}^{0}$, and a noisy version is used as the inverse-network input. During inverse training and validation, independent Gaussian noise is regenerated for each mini-batch with a standard deviation equal to 1\% of the mean absolute head over that batch.

For a fixed candidate interpretation $C_{k}$, let
$\mathcal{B}$ denote the set of indices of the $B$
training examples in a mini-batch. For each example
$i\in\mathcal{B}$, the inverse network predicts the
complete latent vector
$\hat{\mathbf{z}}_{ik}=f_{k}(\mathbf{d}_{i})
\in\mathbb{R}^{d_{z}}$.
Here, $i$ indexes training examples and $k$ indexes
interpretations; neither index denotes a component
of the latent vector. The corresponding predicted
log-conductivity field is
$\hat{\mathbf{Y}}_{ik}
=\mathcal{M}[D_{k}(\hat{\mathbf{z}}_{ik})]$.
The inverse-network loss is
\begin{linenomath*}
\begin{equation}
\begin{aligned}
\mathcal{L}_{\mathrm{inv}}
&=\frac{1}{B}\sum_{i\in\mathcal{B}}
\Bigl[
\lambda_{Y}\operatorname{MSE}
(\hat{\mathbf{Y}}_{ik},\mathbf{Y}_{i}) \\
&\qquad
+\lambda_{z}\operatorname{MSE}
(\hat{\mathbf{z}}_{ik},
 \boldsymbol{\mu}_{k}(\mathbf{x}_{i})) \\
&\qquad
+\lambda_{r}\lVert\hat{\mathbf{z}}_{ik}\rVert_{2}^{2}
\Bigr].
\end{aligned}
\label{eq:inv}
\end{equation}
\end{linenomath*}

Here, $\mathbf{x}_{i}$ is the training image,
$\mathbf{Y}_{i}$ is its target log-conductivity field,
and $\boldsymbol{\mu}_{k}(\mathbf{x}_{i})$ is the
latent mean returned by the frozen VAE encoder.
Each MSE averages over all elements of its arguments
for one training example, and the outer sum averages
the loss over the mini-batch.

The first term penalizes errors in the reconstructed
log-conductivity field, the second penalizes deviations
from the encoder mean, and the third penalizes the
squared Euclidean norm of the complete predicted latent
vector. We use $\lambda_{Y}=1$, $\lambda_{z}=1$, and
$\lambda_{r}=10^{-2}$. Only the inverse-network
parameters are updated; gradients pass through the
frozen decoder to the latent estimate but do not
update the VAE.


For each interpretation, 1550 images are used for training and 50 for validation. At each of 2000 training cycles, 500 training images are sampled without replacement and divided into 15
mini-batches of 32 images and one final mini-batch of
20 images. Optimization uses Adam with a learning rate of $10^{-4}$. After each cycle, the loss in
Equation~\ref{eq:inv} is evaluated over all 50 validation images, and the checkpoint with the lowest mean validation loss is retained.

\subsection{Normalized head error and interpretation ranking}

For observation case $i$ and candidate interpretation $C_k$,
we quantify the discrepancy between observed and predicted
hydraulic heads using the root-mean-square error (RMSE),
\begin{linenomath*}
\begin{equation}
r_{ik}
=
\left[
\frac{1}{N_{\mathrm{obs}}}
\sum_{m=1}^{N_{\mathrm{obs}}}
\left(d_{im}-\hat{d}_{ikm}\right)^2
\right]^{1/2},
\label{eq:rmse}
\end{equation}
\end{linenomath*}
where $N_{\mathrm{obs}}$ is the number of monitoring locations,
and $d_{im}$ and $\hat{d}_{ikm}$ denote the observed and
predicted heads, respectively, at location $m$.
We evaluate all candidate interpretations against the same
observation vector for each case.

We normalize the RMSE by the prescribed observation-noise
standard deviation, $\sigma_{\eta,i}$, to obtain the
dimensionless error
\begin{linenomath*}
\begin{equation}
e_{ik}
=
\frac{r_{ik}}{\sigma_{\eta,i}}.
\label{eq:normalized_error}
\end{equation}
\end{linenomath*}
For the synthetic benchmark, we calculate $\sigma_{\eta,i}$
using Equation~\ref{eq:observation_noise}; Section~4.3
specifies the noise scale for the WIPP application.
We use the same observation-noise scale for all candidates
within a case. Lower values of $e_{ik}$ indicate closer
agreement with the observations relative to the assumed
noise level.

For $N$ evaluation cases, we calculate the aggregate
normalized head error as
\begin{linenomath*}
\begin{equation}
R_k
=
\left[
\frac{1}{N}\sum_{i=1}^{N}e_{ik}^{2}
\right]^{1/2},
\label{eq:aggregate_error}
\end{equation}
\end{linenomath*}
and rank the interpretations by increasing $R_k$.
Each case contributes equally to the mean squared
normalized error. This criterion summarizes performance
across the evaluation set while allowing individual cases
to favor different interpretations. Values of $R_k$
greater than unity indicate that the aggregate head
mismatch exceeds the prescribed observation-noise scale.

For the synthetic benchmark, we also quantify
conductivity-field recovery using the relative $L_2$
error. For test case $i$ and candidate interpretation
$C_k$, let $\mathbf{K}_i^{\mathrm{ref}}$ denote the
reference conductivity field and $\hat{\mathbf{K}}_{ik}$
the recovered field. We calculate
\begin{linenomath*}
\begin{equation}
\varepsilon_{ik}^{(K)}
=
\frac{
\left\|\hat{\mathbf{K}}_{ik}
-\mathbf{K}_i^{\mathrm{ref}}\right\|_2
}{
\left\|\mathbf{K}_i^{\mathrm{ref}}\right\|_2
},
\label{eq:conductivity_diagnostic}
\end{equation}
\end{linenomath*}
where $\|\cdot\|_2$ denotes the Euclidean norm over all
flow-grid cells. We obtain the conductivity fields from
the natural log-conductivity fields as
$\hat{\mathbf{K}}_{ik}=\exp(\hat{\mathbf{Y}}_{ik})$ and
$\mathbf{K}_i^{\mathrm{ref}}=\exp(\mathbf{Y}_i^{\mathrm{ref}})$,
with exponentiation applied elementwise. Because this
metric requires the reference conductivity field, we use
it only to diagnose synthetic-field recovery. It does not
contribute to the interpretation ranking.

\section{Synthetic benchmark}

\subsection{Interpretation classes and generated image ensembles}

We constructed a controlled benchmark based on the Johansen Formation, a candidate geological CO$_2$-storage unit offshore Norway for which geological and flow models have been published \cite{eigestad2009}. We constructed the synthetic reference fields from two-dimensional slices of the original three-dimensional
model. We cropped each slice to remove surrounding white
space while preserving the geologic structures within
the retained region. These cropped slices provided the
reference geometry for the benchmark and the basis for
the property fields used to generate synthetic head
observations and evaluate field recovery.The benchmark uses three written geologic interpretations that differ in their consistency with the reference representation:

\begin{enumerate}
\item \emph{Precise \& Accurate} ($C_1$), which closely represents the reference unit ordering, structural geometry, continuity, and contacts;
\item \emph{Accurate} ($C_2$), which preserves the general geologic setting and principal structure but omits or simplifies features that influence connectivity; and
\item \emph{Mismatched} ($C_3$), which describes a substantially different, predominantly horizontal architecture that is inconsistent with the reference structure.
\end{enumerate}

~\ref{app:image_generation} provides the complete
prompts for the three interpretations and details the
procedure used to generate and prepare the image ensembles.

For each interpretation, the text-conditioned foundation model generated an ensemble of 1600 grayscale images. A separate VAE was trained for each 1600-image ensemble, and a separate inverse network was then trained in the corresponding latent space, following Sections~2.2--2.5. Consequently, each candidate pipeline can produce only conductivity structures represented within its own generated ensemble. The three pipelines were evaluated using the same 595 synthetic reference fields and the same set of 250 monitoring locations.

\begin{figure}
\centering
\includegraphics[width=0.31\textwidth]{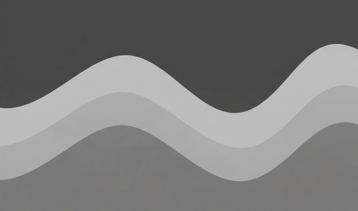}\hfill
\includegraphics[width=0.31\textwidth]{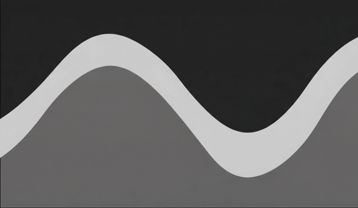}\hfill
\includegraphics[width=0.32\textwidth]{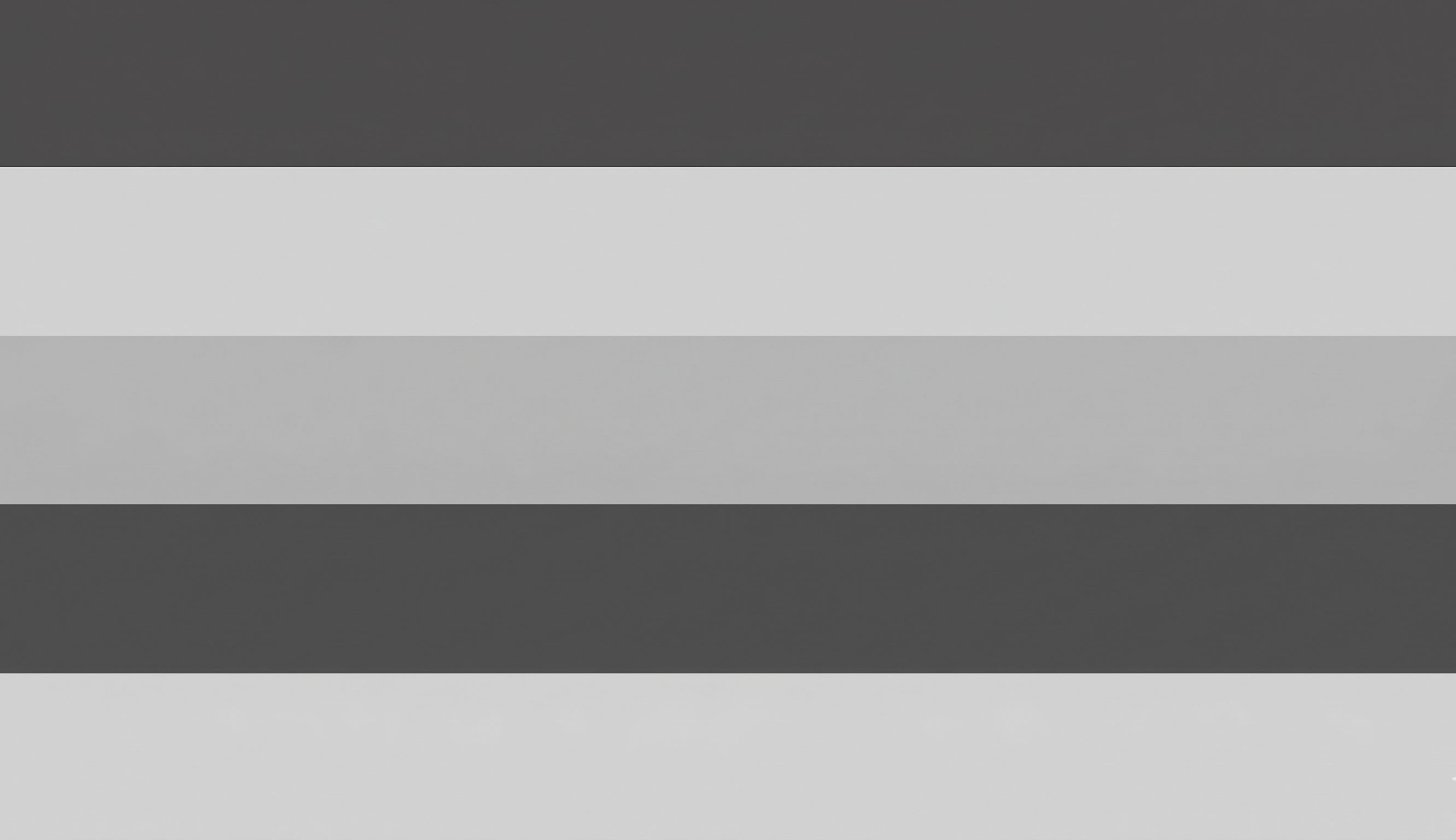}
\caption{Representative images from the three interpretation-specific ensembles. Left: \emph{Precise \& Accurate}. Center: \emph{Accurate}. Right: \emph{Mismatched}. Each ensemble contains 1600 generated images. Brighter intensities are mapped to larger natural log-conductivity values during flow-model construction.}
\label{fig:classes}
\end{figure}

Figure~\ref{fig:classes} presents one generated image from each interpretation-specific ensemble. The \emph{Precise \& Accurate} and \emph{Accurate} ensembles both contain an inclined or folded conductive unit, although the latter represents its internal and bounding geometry less completely. In contrast, the \emph{Mismatched} ensemble is characterized primarily by subhorizontal layering. These differences establish three distinct geologic priors before any head observations are introduced.

\subsection{Representative recovered conductivity fields and head responses}
\begin{figure}[h!]
\centering
\includegraphics[width=\textwidth]
{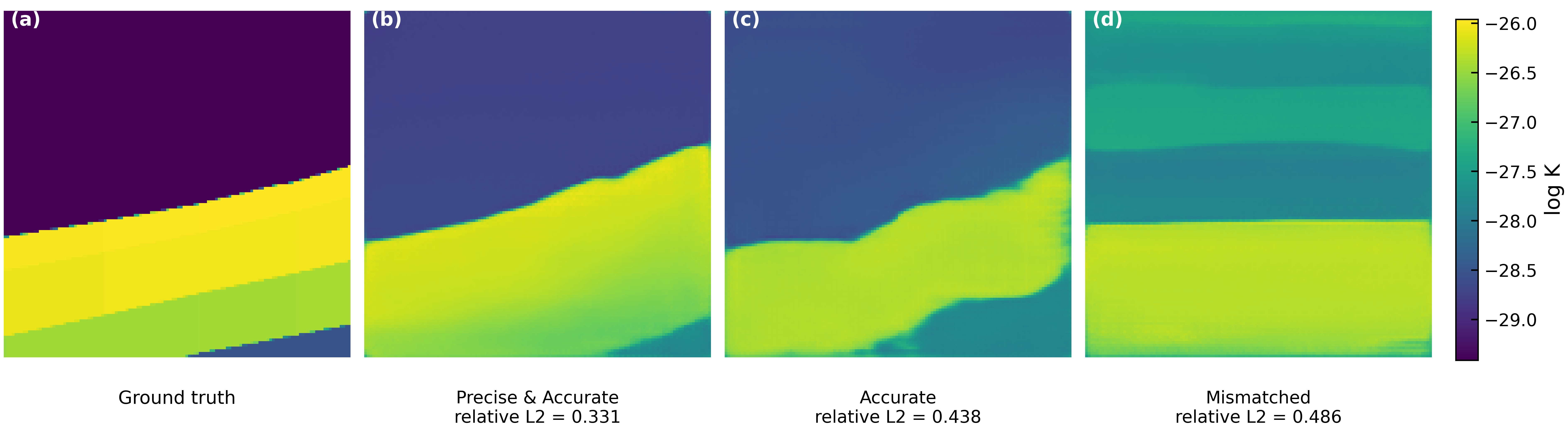}

\vspace{1ex}

\includegraphics[width=\textwidth]
{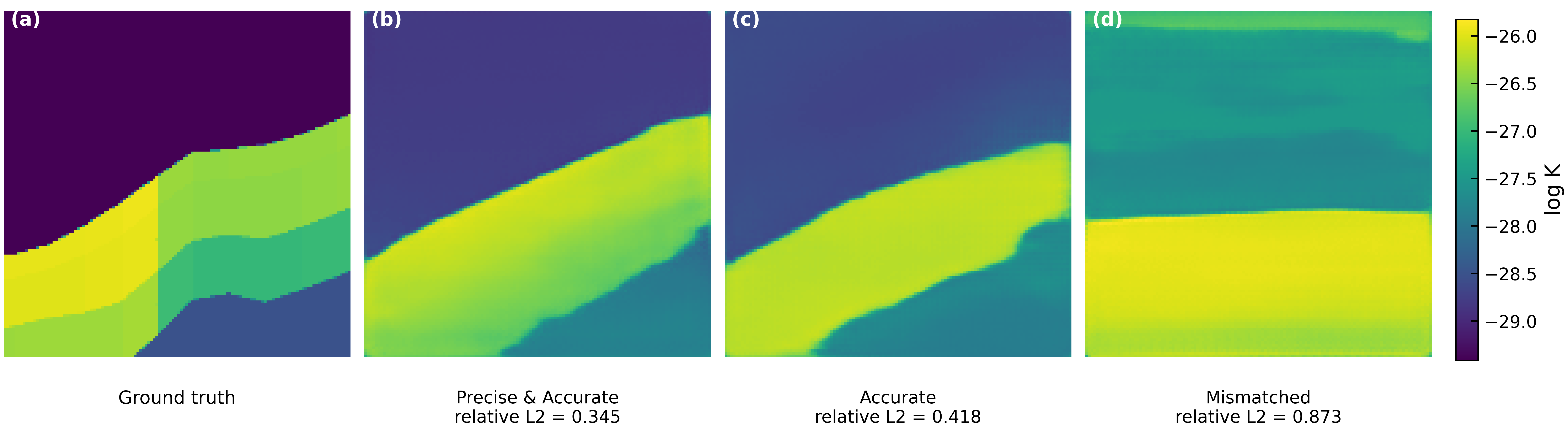}
\caption{Reference and recovered conductivity fields for
test case 1 (top row) and test case 2 (bottom row).
In each row, columns show (a) the reference field,
(b) the \emph{Precise \& Accurate} reconstruction,
(c) the \emph{Accurate} reconstruction, and
(d) the \emph{Mismatched} reconstruction.
Colors represent natural log conductivity, $Y=\ln K$.
The relative $L_2$ errors below the reconstructed fields
are calculated on conductivity, $K=\exp(Y)$, and provide
diagnostics of field recovery rather than inputs to
the interpretation ranking.}
\label{fig:perm_cases}
\end{figure}

\begin{figure}[h!]
\centering
\begin{minipage}[t]{0.49\textwidth}
\centering
\includegraphics[width=\linewidth]
{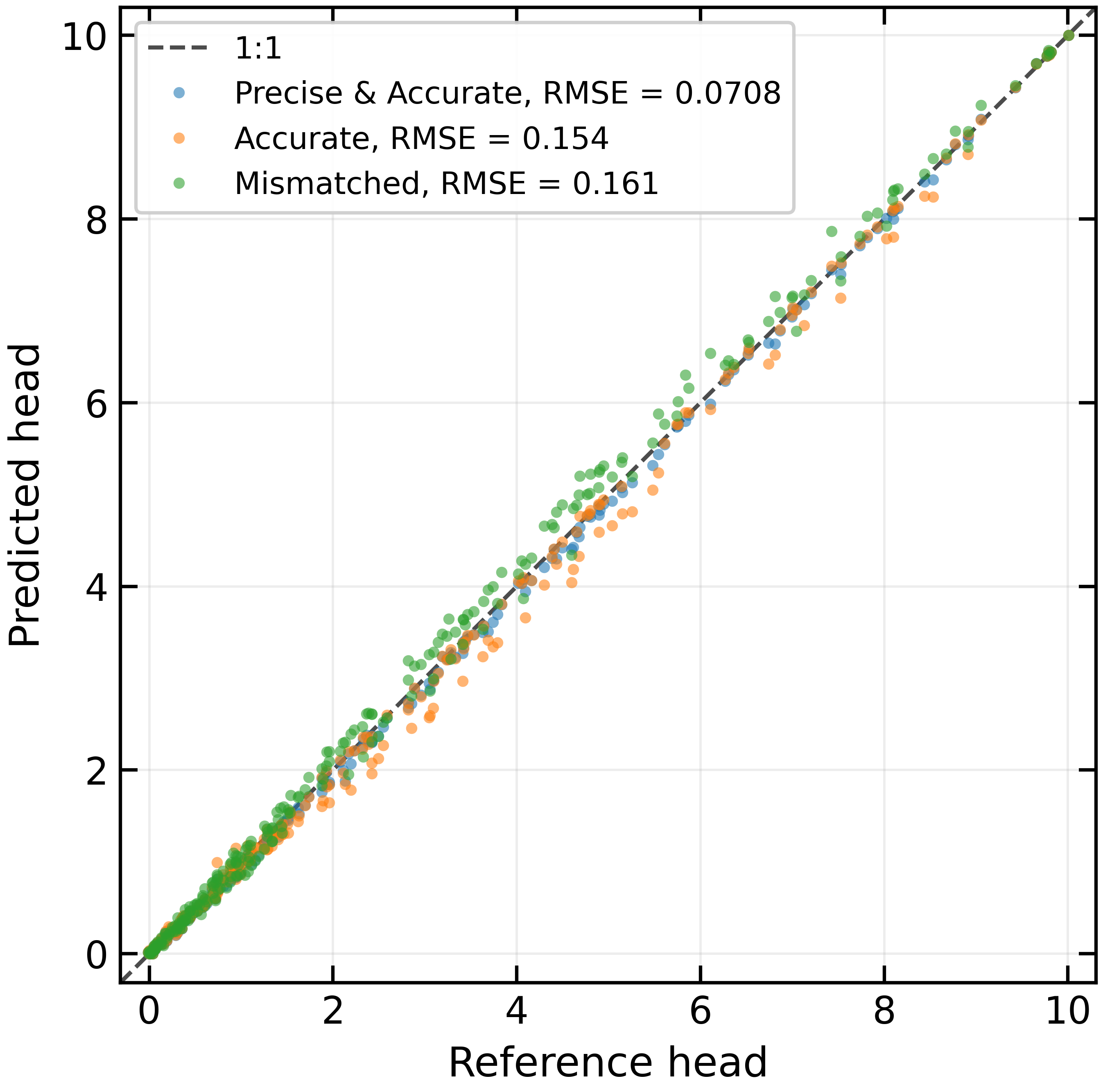}

\small (a) Test case 1
\end{minipage}
\hfill
\begin{minipage}[t]{0.49\textwidth}
\centering
\includegraphics[width=\linewidth]
{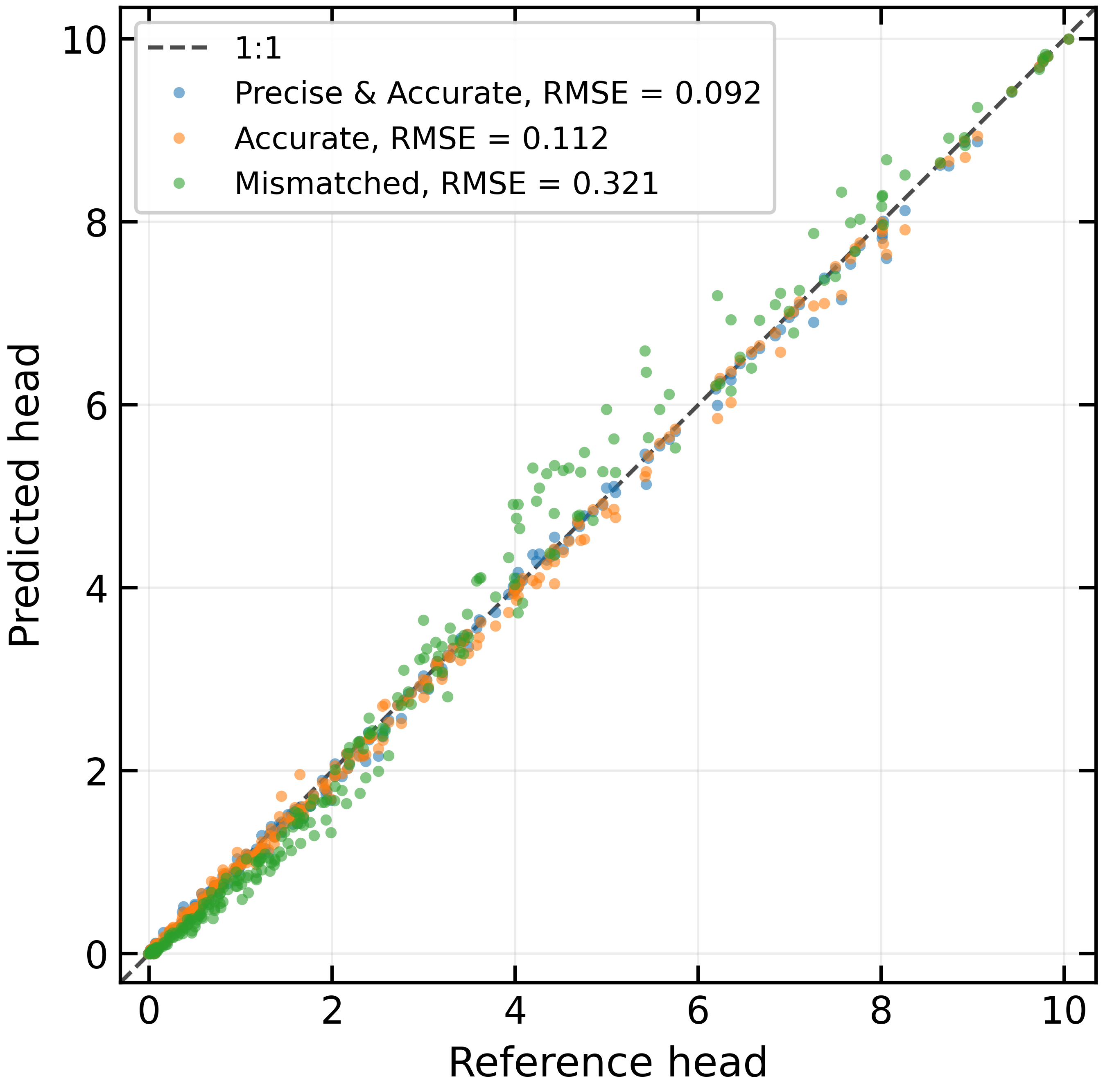}

\small (b) Test case 2
\end{minipage}
\caption{Predicted versus reference hydraulic heads at
the 250 monitoring locations for (a) test case 1 and
(b) test case 2. Each prediction results from forward
simulation of the corresponding recovered conductivity
field in Figure~\ref{fig:perm_cases}.
Dashed lines indicate exact agreement, and the legends
report head RMSEs. The \emph{Precise \& Accurate}
interpretation gives the smallest head misfit in both
cases. The \emph{Accurate} and \emph{Mismatched}
predictions have similar RMSEs in case 1, whereas the
\emph{Mismatched} prediction has a larger misfit in
case 2.}
\label{fig:head_cases}
\end{figure}

We present two representative test cases to compare the recovered conductivity structures and their associated hydraulic-head responses. For each case, we supply the corresponding noisy head-observation vector to all three interpretation-specific inverse networks. Figure~\ref{fig:perm_cases} compares the recovered conductivity fields with the reference fields, and Figure~\ref{fig:head_cases} compares their simulated heads with the reference observations.

The conductivity field comparisons provide a diagnostic of spatial
recovery. The figures show the natural log conductivity, $Y=\ln K$,
and report the calculated relative $L_2$ errors on conductivity,
$K=\exp(Y)$, using Equation~\ref{eq:conductivity_diagnostic}.

In test case 1 (Figure~\ref{fig:perm_cases}, top row), the left most image shows the reference conductivity field. The
\emph{Precise \& Accurate} reconstruction captures its
position and overall inclination most closely, with a
relative conductivity error of 0.331. The
\emph{Accurate} reconstruction retains the inclined unit
but distorts its contacts, yielding an error of 0.438.
The \emph{Mismatched} reconstruction produces
subhorizontal layers and has an error of 0.486.

Test case 2 contains a more pronounced lateral change
in the reference-unit geometry
(Figure~\ref{fig:perm_cases}, bottom row).
The \emph{Precise \& Accurate} and \emph{Accurate}
reconstructions recover the overall inclination but
smooth finer geometric details, with relative
conductivity errors of 0.345 and 0.418, respectively.
The \emph{Mismatched} reconstruction again produces
subhorizontal layering, yielding a substantially larger
error of 0.873. The field diagnostics therefore favor
the \emph{Precise \& Accurate} interpretation in both
examples, although the magnitude of the differences
varies between cases.

The head comparisons show how these differences in
recovered structure affect the simulated observations
(Figure~\ref{fig:head_cases}). In test case 1, the
\emph{Precise \& Accurate} prediction follows the
1:1 line most closely and yields a head RMSE of 0.0708.
The \emph{Accurate} and \emph{Mismatched} predictions
have similar RMSEs of 0.154 and 0.161, respectively.
Thus, these two reconstructions produce comparable
head misfits despite their different conductivity
structures.

In test case 2, the \emph{Precise \& Accurate} and
\emph{Accurate} predictions yield head RMSEs of 0.092
and 0.112, respectively. The \emph{Mismatched}
prediction shows larger departures from the 1:1 line, particularly over the middle and upper portions of the head range, and its RMSE increases to 0.321. The separation between the \emph{Mismatched} interpretation and the other two is therefore greater in this case.

Together, these examples show how the framework discriminates competing geological interpretations according to their ability to reproduce the hydraulic observations and underlying geologic structures.
The \emph{Precise \& Accurate} interpretation yields the
smallest conductivity and head misfit in both cases, supporting its greater consistency with actual geologic formation. Although the degree of separation between candidates varies across cases, the framework provides a common, quantitative basis for comparing interpretations using limited head observations, without requiring knowledge of the true conductivity field.

\subsection{Aggregate head misfit and paired interpretation comparisons}
\begin{figure}[h!]
\centering
\includegraphics[width=\textwidth]
{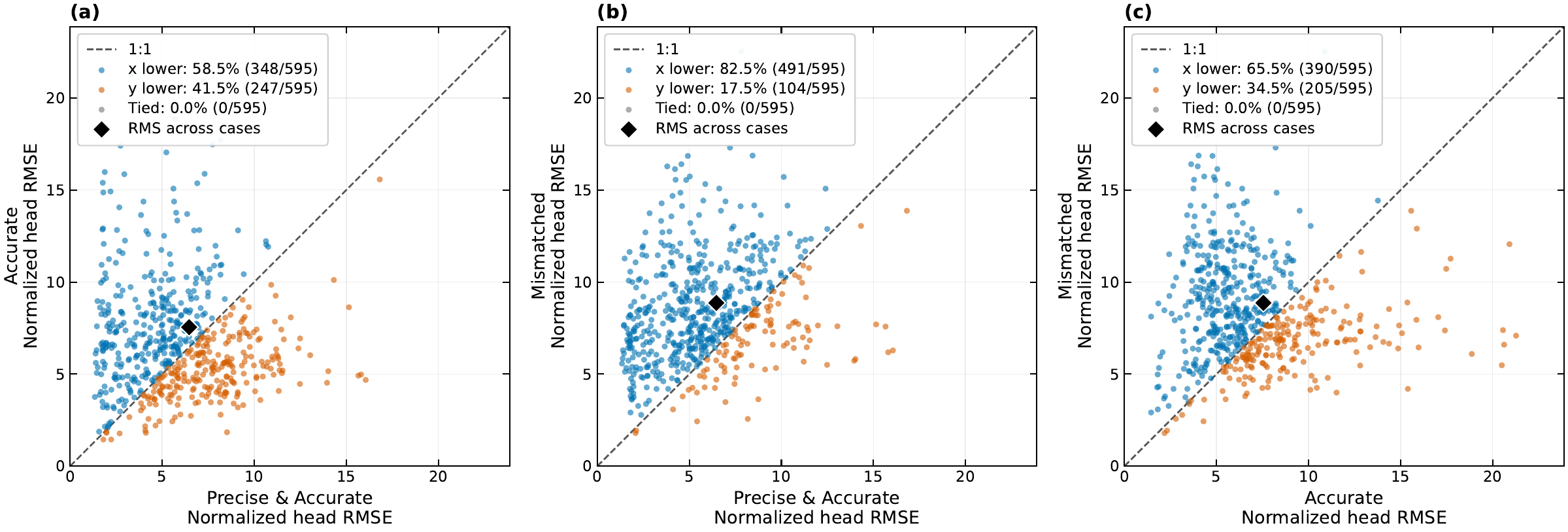}
\caption{Paired comparisons of normalized hydraulic-head
RMSE across 595 synthetic test cases:
(a) \emph{Precise \& Accurate} versus \emph{Accurate},
(b) \emph{Precise \& Accurate} versus \emph{Mismatched},
and (c) \emph{Accurate} versus \emph{Mismatched}.
Each point compares two interpretations for the same
test case. Dashed lines indicate equal normalized error.
Blue points above the lines favor the interpretation
on the horizontal axis; orange points below the lines
favor the interpretation on the vertical axis.
Legends report the corresponding counts and percentages.
Black diamonds show the paired root-mean-square
normalized errors across all cases, corresponding
to the aggregate scores $R_k$ defined in
Equation~\ref{eq:aggregate_error}.}
\label{fig:rmse}
\end{figure}

Figure~\ref{fig:rmse} compares the normalized head errors
of the three interpretations across 595 synthetic test
cases. Each case-level error combines residuals at the
250 monitoring locations and uses the normalization
defined in Section~2.6. The black diamonds summarize
the root-mean-square normalized errors across the
complete test set, corresponding directly to the
aggregate ranking scores $R_k$ in
Equation~\ref{eq:aggregate_error}.

The aggregate scores recover the expected ordering:
\emph{Precise \& Accurate} has the smallest $R_k$,
followed by \emph{Accurate} and then \emph{Mismatched}.
All three black diamonds lie above their respective
1:1 lines, indicating that the interpretation on the
horizontal axis has the lower aggregate error in
each comparison. The ranking therefore favors the
more faithful geological interpretations based on
their overall agreement with the hydraulic
observations. The conductivity diagnostics in
Figure~\ref{fig:perm_cases} provide a separate check
of field recovery and do not contribute to this ranking.

The paired points show how often individual cases
follow the aggregate ordering.
\emph{Precise \& Accurate} produces a lower normalized
error than \emph{Accurate} in 348 cases
(58.5\%; Figure~\ref{fig:rmse}a) and a lower error
than \emph{Mismatched} in 491 cases
(82.5\%; Figure~\ref{fig:rmse}b).
\emph{Accurate} outperforms \emph{Mismatched}
in 390 cases (65.5\%; Figure~\ref{fig:rmse}c).
Thus, the aggregate preference for
\emph{Precise \& Accurate} holds more frequently
against \emph{Mismatched} than against
\emph{Accurate}.

Points below the 1:1 lines identify cases that reverse
the aggregate ordering. These reversals show that
no interpretation provides the smallest error for
every reference field. Nevertheless, aggregating
the normalized errors yields the expected benchmark
ranking from hydraulic observations alone.
The paired comparisons complement that ranking by
quantifying how consistently it holds across cases;
the fractions of cases with lower error are
descriptive frequencies, not the compatibility
weights $W_k$.

\section{Application to the Culebra Dolomite at WIPP}

\subsection{Site and competing conceptual models}

The Waste Isolation Pilot Plant is a geologic repository for transuranic waste, constructed 655\,m below ground in bedded halite of the Permian Salado Formation in southeastern New Mexico. The Culebra Dolomite Member of the overlying Rustler Formation is the most transmissive saturated unit above the repository horizon and the most likely groundwater pathway for radionuclides released by inadvertent human intrusion, which makes its conceptual model consequential for the compliance case \cite{beauheim2008}.

WIPP is useful here because the conceptual model was revised over time. The model supporting the original 1996 license application treated all units below the water-table aquifer as effectively isolated from surface hydrologic processes, with heads slowly declining since the end of the last glacial pluvial period roughly 14{,}000 years ago, and the Culebra as a fully confined unit whose heads would appear steady over the operational period. Subsequent monitoring contradicted this. Culebra heads were found to be rising and to respond to discrete present-day events including major rainstorms, and recalibration failed to reproduce the high-transmissivity offsite pathway the original model placed in the southeastern part of the site. The revised model holds that strata overlying the Culebra in Nash Draw have lost their effectiveness as confining beds, so that heads there respond to rainfall and those changes propagate east into the confined site area. It also treats the region east of the site, where halite occupies Culebra pore space, as having very low transmissivity and very high head \cite{beauheim2008}.

These two accounts are exactly the kind of input the workflow takes. They are prose, they disagree about connectivity and confinement, and there is independent evidence about which is better supported.

\subsection{Conditioning on a boundary outline}

For application to Culebra, the geologic boundary relevant to the comparison is too intricate to specify in words, and images generated from text alone placed structure in the wrong locations. We therefore supplied a contour outline of the domain boundary as an additional conditioning input, so that the foundation model controls the internal architecture while the outline fixes the geometry. Geologic contour maps such as this and other image-based information are often available and can be easily incorporated into our workflow.

\begin{figure}[!h]
\centering
\includegraphics[width=0.30\textwidth]{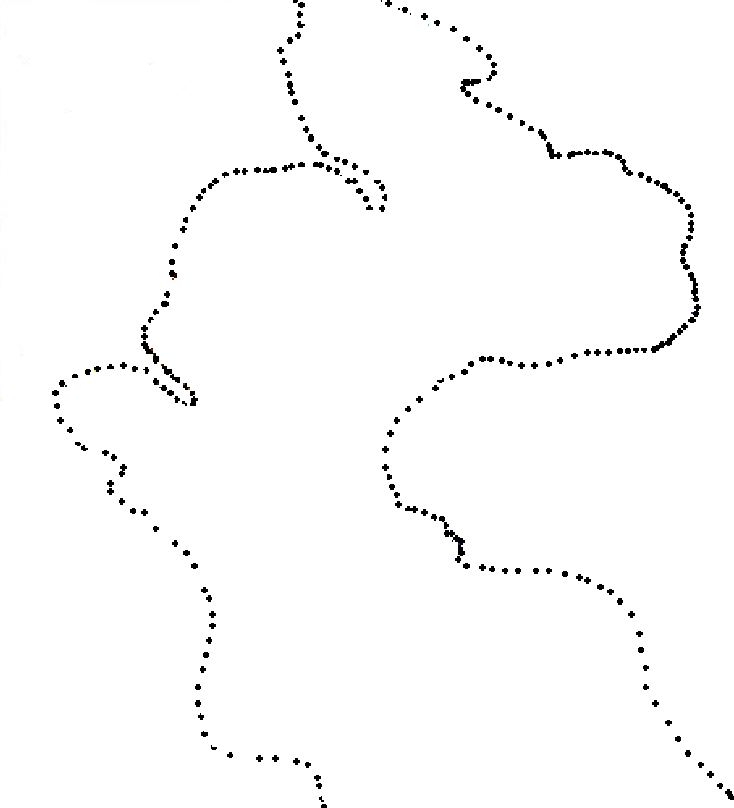}\hfill
\includegraphics[width=0.30\textwidth]{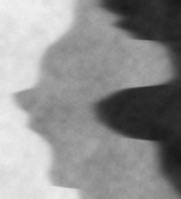}\hfill
\includegraphics[width=0.30\textwidth]{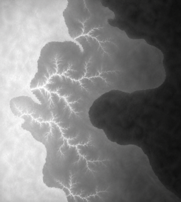}
\caption{Conditioning and generated fields for the Culebra comparison. Left: domain boundary outline supplied to the foundation model alongside the written description. Center: realization generated from the original conceptual model. Right: realization generated from the revised conceptual model.}
\label{fig:wipp}
\end{figure}

Figure~\ref{fig:wipp} shows the outline and one generated realization for each conceptual model. ~\ref{app:image_generation} provides the complete prompts for the two conceptual models and details the
procedure used to generate the image ensembles. The field generated from the original description is smooth and broadly zoned, consistent with a confined unit whose properties vary gradually. The field generated from the revised description contains a branching high-conductivity network, consistent with the fracture connectivity and recharge pathways the revision introduced.

\subsection{Head observations and numerical setup}
\label{sec:wipp_setup}

We use 32 steady-state freshwater-head observations from
the 2000 measurement period, compiled from Table~3 of
\citeA{mckennahart2003}. The heads range from $911.57$ to
$940.03\,\mathrm{m}$. We associate each well with the nearest
flow-grid node using its easting and northing coordinates
and evaluate both conceptual models against the same
observations.

We simulate two-dimensional, steady-state flow on a
rectangular grid with 224 nodes in the east--west direction
and 307 in the north--south direction, giving 68,768 nodes.
The domain-coordinate file defines the horizontal extent.
The model represents spatially variable transmissivity
and contains no internal sources or sinks.

We prescribe heads along the northern, eastern, and southern
boundaries and impose no flow along the western boundary.
We obtain the prescribed heads from a least-squares planar
fit to the observed heads and limit the boundary values
to the observed range extended by $5\,\mathrm{m}$ at each
end. Both conceptual models use the same grid, monitoring
locations, and boundary conditions.

For the original and revised conceptual models, we follow
the image-generation, VAE, and inverse-modeling workflow
described in Section~2. Each interpretation has its own
image ensemble, VAE, and inverse network. For WIPP, the
decoded images map to transmissivity fields, which we
use to predict heads and rank the interpretations.

\subsection{Ranking of the WIPP conceptual models}

We rank the original and revised conceptual models using the aggregate normalized head error, $R_k$, defined in Equation~\ref{eq:aggregate_error}. We evaluate both models over 100 realizations of the inverse-network input. The first realization uses the observed heads directly; the remaining 99 add independent, zero-mean Gaussian perturbations with a standard deviation of $0.5\,\mathrm{m}$. Both models receive the same input vector for each realization, and we calculate all prediction errors against the same unperturbed observations. We normalize the head RMSE by the prescribed observation-noise standard deviation, $\sigma_{\eta,i}=0.5\,\mathrm{m}$, for every realization. These evaluations therefore characterize sensitivity to input perturbations rather than variation among independent field datasets.

\begin{figure}[htbp]
\centering
\includegraphics[width=\textwidth]{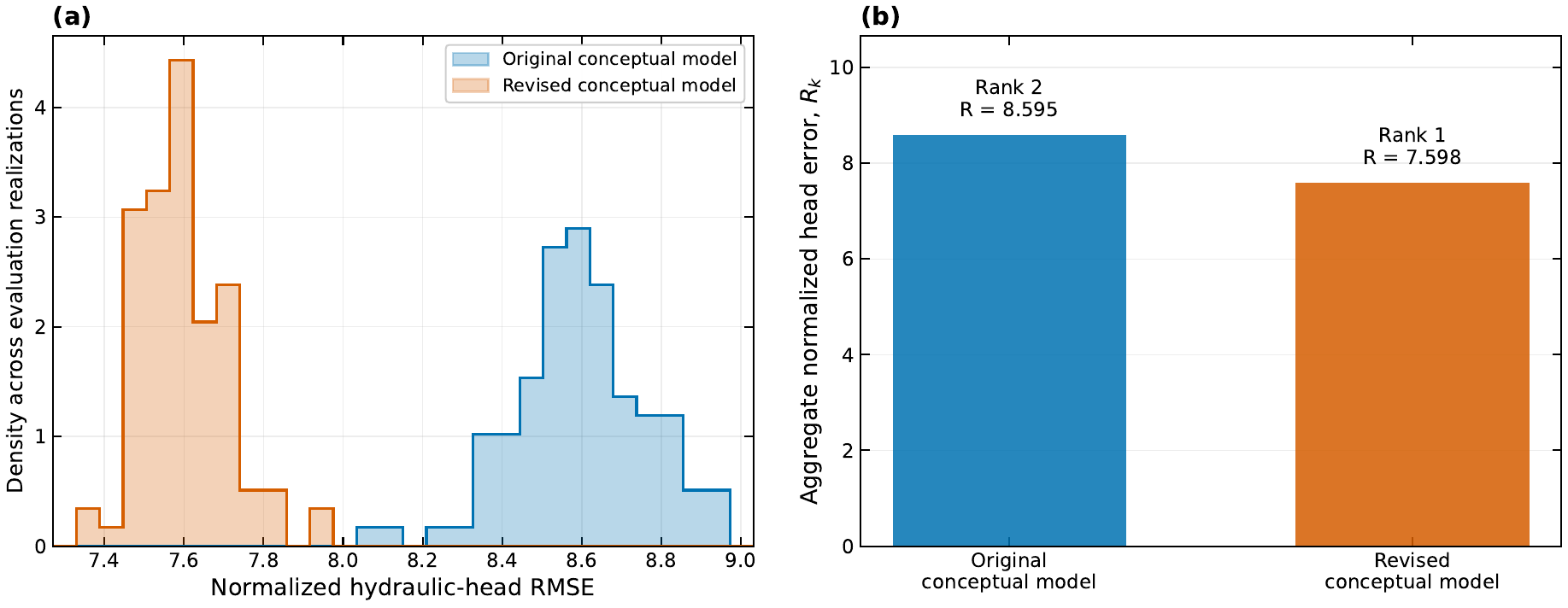}
\caption{Hydraulic-head errors and aggregate ranking of the WIPP conceptual models. (a) Distributions of normalized head RMSE across 100 evaluation realizations: one uses the unperturbed observations as inverse-network input, and 99 use perturbed inputs. Both models receive identical inputs for each realization, and all predictions are scored against the same unperturbed observations using an observation-noise standard deviation of $0.5\,\mathrm{m}$. (b) Aggregate normalized head error, $R_k$, calculated as the root mean square of the realization-specific normalized errors. The revised conceptual model ranks first ($R_k=7.598$), followed by the original conceptual model ($R_k=8.595$).}
\label{fig:wipphead}
\end{figure}

Figure~\ref{fig:wipphead}b presents the resulting aggregate ranking. The revised conceptual model ranks first with $R_k=7.598$, compared with $R_k=8.595$ for the original model, corresponding to an approximately $11.6\%$ reduction in aggregate normalized error. The distributions in Figure~\ref{fig:wipphead}a support this ordering: the revised model produces lower normalized head errors throughout the evaluated ensemble, with no overlap between the two distributions. Thus, the preference for the revised model persists across the prescribed input perturbations.

This ranking is consistent with the site evidence that motivated the conceptual-model revision. The framework favors the revised interpretation because its associated spatial representation produces closer agreement with the observed heads under the common modeling assumptions. However, both aggregate errors substantially exceed unity, indicating that the remaining head mismatch exceeds the assumed observation-noise scale. The results therefore support the revised model relative to the original while also identifying residual discrepancies that neither candidate resolves.

\section{Discussion}

\subsection{What the ranking establishes}

The framework ranks competing geological interpretations according to how closely their inferred property fields reproduce the available hydraulic-head observations. The aggregate normalized error, $R_k$, defined in Equation~\ref{eq:aggregate_error}, provides a common criterion for this comparison: lower values indicate closer agreement relative to the prescribed observation-noise scale. This ranking identifies the best-performing candidate under the adopted modeling assumptions; it does not establish that the interpretation uniquely or completely represents the subsurface. The comparison also remains limited to the supplied candidates. If all candidates omit important geological features, one will still have the smallest error, so the magnitude of the remaining mismatch matters alongside the ordering.

The synthetic benchmark illustrates the value of aggregation. Across the 595 test fields, the \emph{Precise \& Accurate} interpretation has the lowest aggregate normalized error, followed by \emph{Accurate} and \emph{Mismatched}. The pairwise comparisons in Figure~\ref{fig:rmse} nevertheless show reversals in this ordering for individual cases. The aggregate ranking therefore reflects differences across the complete benchmark rather than consistent separation in every case. When alternative structures produce similar heads at the monitoring locations, those observations provide limited information for distinguishing the interpretations. Additional observations or different hydraulic conditions may be needed to resolve their differences.

The WIPP application demonstrates how the framework can use existing monitoring records to compare geological interpretations. The revised conceptual model produces a smaller aggregate normalized error than the original, and this preference persists across the evaluated input perturbations (Figure~\ref{fig:wipphead}). These evaluations assess sensitivity to perturbations of one observed head dataset; they do not represent independent field datasets. The practical contribution is the connection between written interpretations, generated spatial ensembles, and a common hydraulic comparison, which makes alternative geological descriptions easier to evaluate quantitatively.

\subsection{Relation to other machine learning approaches}

Generative models already provide low-dimensional representations of geological heterogeneity for inverse modeling \cite{laloy2017,laloy2018,mo2020}. These methods learn spatial patterns from specified training data and constrain inferred fields through the resulting representation. Our workflow builds on this approach by generating a separate training ensemble from each written interpretation. The main contribution is therefore the connection between geological descriptions and interpretation-specific representations that can be compared using the same observations and scoring criterion.

The workflow also complements approaches that use language-model agents to automate numerical modeling. For example, \citeA{ma2026} developed agents that generate, debug, and adapt calibration code for process-based simulations. Their emphasis is on automating implementation and calibration, whereas our emphasis is on evaluating alternative geological representations. A future workflow could combine these capabilities by using agents to manage simulation and calibration consistently across candidate interpretations while retaining explicit control over the geological assumptions being compared.

\subsection{Limitations}

The generated images represent spatial patterns rather than calibrated hydraulic properties. Their grayscale values describe relative contrasts and features such as layering, contacts, and connectivity. The mapping from image intensity to conductivity or transmissivity requires separately specified property ranges. Consequently, the ranking depends on both the represented geometry and the adopted property mapping. Independent geological and hydraulic information remains necessary to constrain these choices.

The comparison also evaluates the complete inference pipeline associated with each interpretation. A larger head mismatch may reflect an unsuitable geological representation, limited decoder expressiveness, imperfect inverse-network training, or forward-model assumptions. The present comparison does not isolate these contributions. Repeated training runs and additional optimization within each latent space would help determine whether the ordering persists when inference errors are reduced. Because the same observations condition the inverse solution and enter the scoring calculation, the reported errors measure consistency with those observations. Evaluation against withheld observations would provide a separate test of predictive performance.

The forward models used here describe steady-state, single-phase flow in two dimensions. They cannot evaluate differences that become apparent only through transient responses, multiphase flow, or three-dimensional connectivity. Extending the framework to these settings would require an appropriate forward solver, new simulation-based training data, and retraining of the inverse networks. Three-dimensional applications would also require a suitable representation of volumetric geological structure.

Finally, the generated ensembles depend on prompt wording, auxiliary images, and the image-generation model. Alternative descriptions of the same interpretation may produce different spatial patterns, and we have not quantified the resulting variation in ranking. Outline images can help convey known geometry, as in the WIPP application, but they do not ensure that every generated realization preserves the intended geological relationships. Future evaluation should examine sensitivity to prompt wording, independently generated ensembles, and model versions. Improvements in image generation may increase fidelity to the supplied descriptions, but their effect on geological realism and interpretation ranking must be assessed explicitly.

\section{Conclusions}

We developed a framework that connects written geological interpretations to quantitative evaluation using sparse hydraulic-head observations. A text-conditioned image-generation model produces an ensemble for each interpretation, a separate VAE learns its spatial representation, and an inverse network maps observations into the corresponding latent space. Forward simulation then provides head predictions, and the aggregate normalized error ranks the candidate interpretations.

Across 595 synthetic test fields, the framework recovers the expected aggregate ordering: \emph{Precise \& Accurate}, followed by \emph{Accurate} and \emph{Mismatched}, despite reversals for individual cases. For the WIPP application, the revised conceptual model ranks first with $R_k=7.598$, compared with $R_k=8.595$ for the original model. This approximately $11.6\%$ reduction is consistent with the direction of the documented conceptual-model revision, although neither candidate reproduces the observations within the assumed noise scale.

These results demonstrate that written geological interpretations can serve as inputs to a systematic comparison using hydraulic observations. Auxiliary images can supplement descriptions when known geometry is difficult to express in words. The framework provides a practical means of identifying which supplied interpretation produces the closest agreement with the data, subject to the generated representation, inverse-model performance, and physical assumptions. Broader application will require evaluating these sensitivities and testing predictions against independent observations.

\appendix
\section{Image-generation procedure and prompts}
\label{app:image_generation}

\subsection{Image-generation procedure}

We generated the geological images using the image-generation
feature in ChatGPT Enterprise (OpenAI), with GPT-5.5 selected
in the conversation interface. The GPT-5.5 designation
identifies the selected conversational model. We supplied
written prompts that describe the spatial characteristics
associated with each interpretation, including unit ordering,
relative property contrasts, structural geometry, and
connectivity.

For the synthetic benchmark, we assembled an ensemble of
1600 grayscale images for each of the three interpretation
classes. The prompts request solid grayscale regions without
labels, grid lines, or blank margins. For the WIPP application,
we additionally supplied the outline shown in
Figure~\ref{fig:wipp} to guide the spatial arrangement of the
zones. Each recorded WIPP prompt requests 100 images of
uniform dimensions, with variation in relative conductivity
and channel geometry.

The following subsections reproduce the recorded prompts,
including their original wording.

\subsection{Synthetic-benchmark prompts}

\subsubsection{Precise \& Accurate interpretation}
\label{app:prompt_precise}

The prompt specifies four units with approximate
thickness fractions of 40\%, 20\%, 15\%, and 25\%, respectively.
It requests an undulating anticlinal geometry and uses
different gray levels to distinguish the units.

\begin{quote}
Create a 2D image of reservoir on a gray scale with four
layers with four different colors. Layer 1 (Top layer):
dark gray (Approx 40 \% of total thickness), which represent
the caprock or low conductivity layer; Layer 2, which is
a conductive reservoir rock layer (Approx., 20\% of total
thickness): light gray, Layer 3: also, a reservoir rock
layer, (approx. 15\% of total thickness): slightly darker
light gray; Layer 4 : a low conductive reservoir layer,
(approx. 25\% of total thickness): medium dark gray.
The layers are undulating, like a curve forming like a
geologic anticline. No grid line, no labels, no white space.
There should be no boundary line between the layers as
well. These four layers should cover the whole image.
Create the image in solid color
\end{quote}

\subsubsection{Accurate interpretation}
\label{app:prompt_accurate}

The prompt simplifies the architecture to a
caprock above two reservoir units. It requests curved
layers but does not specify their relative thicknesses.

\begin{quote}
Create a 2D image of a reservoir on a gray scale with three
layers with three different layers in three different
colors. Layer 1: very dark gray, which is a caprock layer;
Layer 2 light gray which is reservoir layer, Layer 3 medium
dark gray, which is also a reservoir layer. The layers are
like a curve forming a geologic formation. No grid line,
no labels, no white space. These three layers should cover
the whole image. Create the image in solid color.
\end{quote}

\subsubsection{Mismatched interpretation}
\label{app:prompt_mismatched}

The prompt specifies five units, including two caprock layers and three reservoir layers. It leaves layer thicknesses and structural curvature unspecified.

\begin{quote}
Create a geologic image on a gray scale with five horizontal reservoir layers, first a caprock in dark gray then two reservoir layers and then another caprock layer and then another reservoir
layer. No grid line, no labels, no white space. These five
layers should cover the whole image. Create the image
in solid color.
\end{quote}

\subsection{WIPP prompts and outline conditioning}
\label{app:prompt_wipp}

For both WIPP interpretations, we supplied a zone outline shown in Figure \ref{fig:wipp} alongside the written prompt. Both prompts assign high relative conductivity to the leftmost zone, moderate relative conductivity to the middle zone, and low relative conductivity to the rightmost zone. 

Both prompts also request variation in high-conductivity
channels. The revised-model prompt adds a specific
requirement for branched channels that extend from the
high-conductivity zone into the moderate-conductivity zone.
Thus, the explicit difference between the prompts concerns
channel geometry and interzone connectivity.

\subsubsection{Original conceptual model}
\label{app:prompt_wipp_original}

\begin{quote}
Using the attached outline of reservoir zones, the left
most zone has high conductivity, the middle one has
moderate conductivity and the right most one has low
conductivity. create a conductivity map for this reservoir
in gray scale, which means high perm is lightest, moderate
lighter and low conductivity means dark gray. Remove the
outline, the image needs to be high resolution. create
100 images like this bring slight variation in conductivity
and large variation in high perm channels. zip the image,
the image needs to be of same size.
\end{quote}

\subsubsection{Revised conceptual model}
\label{app:prompt_wipp_revised}

\begin{quote}
Using the attached outline of reservoir zones, the left
most zone has high conductivity, the middle one has
moderate conductivity and the right most one has low
conductivity. create a conductivity map for this reservoir
in gray scale, which means high perm is lightest, moderate
lighter and low conductivity means dark gray. Also, in
the middle moderate conductivity zone add some high perm
branched channels, which creates connectivity from the
high to moderate zone. remove the outline, the image
needs to be high resolution. create 100 images like this
bring slight variation in conductivity and large variation
in high perm channels. zip the image, the image needs
to be of same size. The high-perm channels should look
like a natural branched extension from high to moderate zone.
\end{quote}

\section*{Open Research Section}
The DPFEHM differentiable flow simulator used for all forward solves is openly available \cite{omalley2023}. Source code for autoencoder training, latent inversion, and interpretation ranking, together with the generated image ensembles and the scripts that produce every figure in this paper, are archived at https://github.com/hrashid10/GeolEarlyRank . Observed head data for the Culebra are available through the WIPP records described by \citeA{beauheim2008}.


\section*{Conflict of Interest disclosure}
The authors declare there are no conflicts of interest for this manuscript.

\section*{Author Contributions}
\textbf{Conceptualization:} Harun Ur Rashid,
Daniel O’Malley; \textbf{Data curation:} Harun Ur Rashid;
\textbf{Formal analysis:} Harun Ur Rashid, Daniel O’Malley;
\textbf{Funding acquisition:} Harun Ur Rashid
\textbf{Investigation:} Harun Ur Rashid, Daniel O’Malley
\textbf{Methodology:} Harun Ur Rashid
\acknowledgments
HR was supported by the Laboratory Directed Research and Development program at Los Alamos National Laboratory as part of the project "Physics-informed Generative Inversion for Early Ranking of Competing Geologic Interpretations" under award number: 20261470IR. Los Alamos National Laboratory is operated by Triad National Security, LLC, for the National Nuclear Security Administration of the U.S. Department of Energy.
DO was supported by the U.S. Department of Energy, Office of Science, Office of Basic Energy Sciences, Geosciences program under Award Number LANLECA1.

\bibliography{references}

@article{bond2007,
  author  = {Bond, C. E. and Gibbs, A. D. and Shipton, Z. K. and Jones, S.},
  title   = {What do you think this is? ``Conceptual uncertainty'' in geoscience interpretation},
  journal = {GSA Today},
  volume  = {17},
  number  = {11},
  pages   = {4--10},
  year    = {2007},
  doi     = {10.1130/GSAT01711A.1}
}

@article{bredehoeft2005,
  author  = {Bredehoeft, J.},
  title   = {The conceptualization model problem---surprise},
  journal = {Hydrogeology Journal},
  volume  = {13},
  number  = {1},
  pages   = {37--46},
  year    = {2005},
  doi     = {10.1007/s10040-004-0430-5}
}

@article{hojberg2005,
  author  = {H{\o}jberg, A. L. and Refsgaard, J. C.},
  title   = {Model uncertainty---parameter uncertainty versus conceptual models},
  journal = {Water Science and Technology},
  volume  = {52},
  number  = {6},
  pages   = {177--186},
  year    = {2005},
  doi     = {10.2166/wst.2005.0166}
}

@article{refsgaard2012,
  author  = {Refsgaard, J. C. and Christensen, S. and Sonnenborg, T. O. and Seifert, D. and H{\o}jberg, A. L. and Troldborg, L.},
  title   = {Review of strategies for handling geological uncertainty in groundwater flow and transport modeling},
  journal = {Advances in Water Resources},
  volume  = {36},
  pages   = {36--50},
  year    = {2012},
  doi     = {10.1016/j.advwatres.2011.04.006}
}

@inproceedings{kingma2014,
  author    = {Kingma, D. P. and Welling, M.},
  title     = {Auto-encoding variational {B}ayes},
  booktitle = {Proceedings of the 2nd International Conference on Learning Representations (ICLR)},
  year      = {2014},
  note      = {arXiv:1312.6114}
}

@article{ma2026,
  author  = {Ma, F. and Chen, J. and Dai, Z. and Cai, F. and Hu, Y.},
  title   = {Autonomous inverse modeling of complex groundwater systems via a physics-integrated large language model multi-agent framework},
  journal = {Water Research},
  volume  = {299},
  pages   = {125886},
  year    = {2026},
  doi     = {10.1016/j.watres.2026.125886}
}

@article{omalley2023,
  author  = {O'Malley, D. and Greer, S. Y. and Pachalieva, A. and Hao, W. and Harp, D. and Vesselinov, V. V.},
  title   = {{DPFEHM}: a differentiable subsurface physics simulator},
  journal = {Journal of Open Source Software},
  volume  = {8},
  number  = {90},
  pages   = {4560},
  year    = {2023},
  doi     = {10.21105/joss.04560}
}

@article{innes2018,
  author  = {Innes, M.},
  title   = {Flux: Elegant machine learning with {J}ulia},
  journal = {Journal of Open Source Software},
  volume  = {3},
  number  = {25},
  pages   = {602},
  year    = {2018},
  doi     = {10.21105/joss.00602}
}

@article{bezanson2017,
  author  = {Bezanson, J. and Edelman, A. and Karpinski, S. and Shah, V. B.},
  title   = {{J}ulia: A fresh approach to numerical computing},
  journal = {SIAM Review},
  volume  = {59},
  number  = {1},
  pages   = {65--98},
  year    = {2017},
  doi     = {10.1137/141000671}
}

@article{eigestad2009,
  author  = {Eigestad, G. T. and Dahle, H. K. and Hellevang, B. and Riis, F. and Johansen, W. T. and {\O}ian, E.},
  title   = {Geological modeling and simulation of {CO}$_2$ injection in the {J}ohansen formation},
  journal = {Computational Geosciences},
  volume  = {13},
  number  = {4},
  pages   = {435--450},
  year    = {2009},
  doi     = {10.1007/s10596-009-9153-y}
}

@techreport{beauheim2008,
  author      = {Beauheim, R. L.},
  title       = {Collection and integration of geoscience information to revise the {WIPP} hydrology conceptual model},
  institution = {Sandia National Laboratories},
  number      = {SAND2008-3257P},
  address     = {Albuquerque, NM},
  year        = {2008}
}

@book{caers2011,
  author    = {Caers, Jef},
  title     = {Modeling Uncertainty in the Earth Sciences},
  publisher = {John Wiley \& Sons},
  year      = {2011},
  doi       = {10.1002/9781119995920},
  isbn      = {9781119992622}
}

@incollection{wellmann2018,
  author    = {Wellmann, Florian and Caumon, Guillaume},
  title     = {{3-D} Structural Geological Models: Concepts, Methods, and Uncertainties},
  booktitle = {Advances in Geophysics},
  editor    = {Schmelzbach, Cedric},
  publisher = {Elsevier},
  volume    = {59},
  pages     = {1--121},
  year      = {2018},
  doi       = {10.1016/bs.agph.2018.09.001}
}

@article{refsgaard2006,
  author  = {Refsgaard, Jens Christian and van der Sluijs, Jeroen P. and Brown, James and van der Keur, Peter},
  title   = {A Framework for Dealing with Uncertainty Due to Model Structure Error},
  journal = {Advances in Water Resources},
  volume  = {29},
  number  = {11},
  pages   = {1586--1597},
  year    = {2006},
  doi     = {10.1016/j.advwatres.2005.11.013}
}

@article{demyanov2019,
  author  = {Demyanov, Vasily and Arnold, Dan and Rojas, Temistocles and Christie, Mike},
  title   = {Uncertainty Quantification in Reservoir Prediction: Part 2---Handling Uncertainty in the Geological Scenario},
  journal = {Mathematical Geosciences},
  volume  = {51},
  pages   = {241--264},
  year    = {2019},
  doi     = {10.1007/s11004-018-9755-9}
}

@article{li2011,
  author  = {Li, Shuiquan and Zhang, Ye and Zhang, Xu},
  title   = {A Study of Conceptual Model Uncertainty in Large-Scale {CO$_2$} Storage Simulation},
  journal = {Water Resources Research},
  volume  = {47},
  number  = {5},
  pages   = {W05534},
  year    = {2011},
  doi     = {10.1029/2010WR009707}
}

@book{scheidt2018,
  editor    = {Scheidt, C{\'e}line and Li, Lewis and Caers, Jef},
  title     = {Quantifying Uncertainty in Subsurface Systems},
  series    = {Geophysical Monograph Series},
  volume    = {236},
  publisher = {American Geophysical Union and John Wiley \& Sons},
  year      = {2018},
  doi       = {10.1002/9781119325888},
  isbn      = {9781119325833}
}

@article{neuman2003,
  author  = {Neuman, Shlomo P.},
  title   = {Maximum Likelihood {Bayesian} Averaging of Uncertain Model Predictions},
  journal = {Stochastic Environmental Research and Risk Assessment},
  volume  = {17},
  number  = {5},
  pages   = {291--305},
  year    = {2003},
  doi     = {10.1007/s00477-003-0151-7}
}

@article{ye2004,
  author  = {Ye, Ming and Neuman, Shlomo P. and Meyer, Philip D.},
  title   = {Maximum Likelihood {Bayesian} Averaging of Spatial Variability Models in Unsaturated Fractured Tuff},
  journal = {Water Resources Research},
  volume  = {40},
  number  = {5},
  pages   = {W05113},
  year    = {2004},
  doi     = {10.1029/2003WR002557}
}

@article{laloy2017,
  author  = {Laloy, Eric and H{\'e}rault, Romain and Lee, John and Jacques, Diederik and Linde, Niklas},
  title   = {Inversion Using a New Low-Dimensional Representation of Complex Binary Geological Media Based on a Deep Neural Network},
  journal = {Advances in Water Resources},
  volume  = {110},
  pages   = {387--405},
  year    = {2017},
  doi     = {10.1016/j.advwatres.2017.09.029}
}

@article{laloy2018,
  author  = {Laloy, Eric and H{\'e}rault, Romain and Jacques, Diederik and Linde, Niklas},
  title   = {Training-Image-Based Geostatistical Inversion Using a Spatial Generative Adversarial Neural Network},
  journal = {Water Resources Research},
  volume  = {54},
  number  = {1},
  pages   = {381--406},
  year    = {2018},
  doi     = {10.1002/2017WR022148}
}

@article{mo2020,
  author  = {Mo, Shaoxing and Zabaras, Nicholas and Shi, Xiaoqing and Wu, Jichun},
  title   = {Integration of Adversarial Autoencoders with Residual Dense Convolutional Networks for Estimation of Non-{Gaussian} Hydraulic Conductivities},
  journal = {Water Resources Research},
  volume  = {56},
  number  = {2},
  pages   = {e2019WR026082},
  year    = {2020},
  doi     = {10.1029/2019WR026082}
}

@article{strebelle2002,
  author  = {Strebelle, Sebastien},
  title   = {Conditional Simulation of Complex Geological Structures Using Multiple-Point Statistics},
  journal = {Mathematical Geology},
  volume  = {34},
  number  = {1},
  pages   = {1--21},
  year    = {2002},
  doi     = {10.1023/A:1014009426274}
}

@book{mariethoz2014,
  author    = {Mariethoz, Gr{\'e}goire and Caers, Jef},
  title     = {Multiple-Point Geostatistics: Stochastic Modeling with Training Images},
  publisher = {John Wiley \& Sons},
  year      = {2014},
  doi       = {10.1002/9781118662953},
  isbn      = {9781118662755}
}

@inproceedings{rombach2022,
  author    = {Rombach, Robin and Blattmann, Andreas and Lorenz, Dominik and Esser, Patrick and Ommer, Bj{\"o}rn},
  title     = {High-Resolution Image Synthesis with Latent Diffusion Models},
  booktitle = {Proceedings of the IEEE/CVF Conference on Computer Vision and Pattern Recognition},
  pages     = {10684--10695},
  year      = {2022},
  doi       = {10.1109/CVPR52688.2022.01042}
}

@misc{openai2026chatgpt,
  author       = {{OpenAI}},
  title        = {{ChatGPT Enterprise} ({GPT-5.5})},
  year         = {2026},
  howpublished = {Large language model used with the built-in Image Generation tool},
  url          = {https://chatgpt.com/},
  note         = {Accessed [Month Day--Month Day, 2026]}
}

@techreport{mckennahart2003,
  author      = {McKenna, S. A. and Hart, D. B.},
  title       = {Analysis Report: Task 4 of {AP-088}; Conditioning of Base {T} Fields to Transient Heads},
  institution = {Sandia National Laboratories},
  address     = {Carlsbad, NM},
  number      = {ERMS 531124},
  year        = {2003},
  url         = {https://www.wipp.energy.gov/library/cra/2009_cra/references/Others/McKenna_Hart_2003_Analysis_Report_Task_4_AP088_ERMS531124.pdf}
}

\end{document}